\documentclass[10pt, a4paper]{article}

\usepackage{lrec-coling2024} 
\usepackage{multirow}
\usepackage{booktabs}
\usepackage{enumitem}
\usepackage{amsmath}
\usepackage{amsfonts}
\usepackage{subfigure}
\usepackage{colortbl, xcolor}

\title{Reinforcement Retrieval Leveraging Fine-grained Feedback for Fact Checking News Claims with Black-Box LLM}

\name{Xuan Zhang, ~Wei Gao} 

\address{School of Computing and Information Systems, Singapore Management University \\
         80 Stamford Road, Singapore 178902\\
         xuanzhang.2020@phdcs.smu.edu.sg, ~weigao@smu.edu.sg\\}

\begin{document}
\abstract{
Retrieval-augmented language models have exhibited promising performance across various areas of natural language processing (NLP), including fact-critical tasks. However, due to the black-box nature of advanced large language models (LLMs) and the non-retrieval-oriented supervision signal of specific tasks, the training of retrieval model faces significant challenges under the setting of black-box LLM. We propose an approach leveraging Fine-grained Feedback with Reinforcement Retrieval (FFRR) to enhance fact-checking on news claims by using black-box LLM. FFRR adopts a two-level strategy to gather fine-grained feedback from the LLM, which serves as a reward for optimizing the retrieval policy, by rating the retrieved documents based on the non-retrieval ground truth of the task. We evaluate our model on two public datasets for real-world news claim verification, and the results demonstrate that FFRR achieves significant improvements over strong LLM-enabled and non-LLM baselines.
 \\ \newline  }
\maketitleabstract
\pagestyle{empty}
\section{Introduction}

Recent advancements in large language models (LLMs), such as GPT-4~\cite{openai2023gpt4} and Palm~2~\cite{anil2023palm}, have demonstrated impressive capabilities in generating coherent, informative, and fluent verbal reasoning~\cite{wei2022chain,wang2022self,zhou2022least}. 
This opens up the possibility for applying LLMs in fact-checking related tasks, such as predicting claim veracity and generating explainable rationales~\cite{zhao2023verify,yao2023react,pan2023factchecking,bubeck2023sparks,zhang-gao-2023-towards}. 
However, these tasks heavily rely on up-to-date information and reliable knowledge, and LLMs may fall short in these aspects, potentially affecting their downstream performance with bias and hallucination~\cite{openaiblog,ye2022unreliability,bang2023multitask}.

Inspired by retrieval-augmented language modeling approach~\cite{lewis2020retrieval,guu2020retrieval}, recent efforts have attempted to incorporate retrieved external knowledge with LLMs~\cite{nakano2021webgpt,khattab2023demonstratesearchpredict,press2022measuring,yao2023react,peng2023check}. A common idea is to integrate a plug-and-play retrieval model with the LLM for providing relevant context.
However, the quality of retrieval results can compromise the effectiveness of LLM-based reasoning and verification~\cite{yao2023react}. A key unresolved issue is how to tune the retrieval model for returning the ``smoking gun'' evidence that can assist the LLM in determining the veracity of the claim reliably.


\begin{figure}[t!]
\centering
\includegraphics[width=3in]{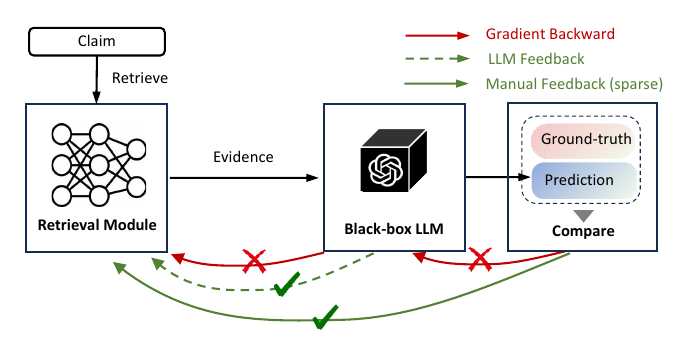}
\caption{The black-box LLM handicaps end-to-end task training through gradient backward. However, manual (sparse) and LLM (fine-grained) feedback can be used to optimize the retrieval model.}
\label{fig:intro}
\end{figure}

Meanwhile, advanced LLMs typically exhibit a black-box nature and often provide service via APIs~\cite{sun2022black}, preventing end-users from accessing their internal mechanisms, such as model parameters and gradients. 
Therefore, it is infeasible to employ the retrieval-augmented approach~\cite{lewis2020retrieval,guu2020retrieval} for end-to-end optimization because it is difficult to backpropagate prediction loss to the retrieval model through the LLM, as depicted in Figure~\ref{fig:intro}. In addition, the ground truth is typically not retrieval-oriented, which may not be used directly for training the retrieval model.

To optimize the retrieval model for adaptation to LLM, 
a recent REtrieve-and-PLUG (REPLUG) approach for open-domain QA employs the LLM to rate retrieved top-$K$ documents with reference to ground truth answers and updates the retriever via minimizing Kullback-Leibler Divergence (KLD) between retrieval score distribution and the rated score distribution ~\cite{shi2023replug}.
From the factuality standpoint, however, such an approach oversimplifies the interaction between the retrieval model and LLM in document selection. Using top-$K$ documents may incorporate unnecessary noise or miss useful evidence, and may fail to account for complex real-world claims. For example, for a claim like ``Bruce Lee died from drinking too much water'', top search results predominantly stem from a report that Lee may have had too much water on the day of his death, and his kidneys lacked the ability to sufficiently shed it, which might not be key evidence\footnote{Fact-checking websites indicate the case is unproven because most researchers believe that Lee had several risk factors for the disease, such as swelling on the brain and heatstroke, and the claim is merely a recent hypothesis put forward by several medical researchers. \url{https://www.snopes.com/fact-check/bruce-lee-die-drinking-too-much-water/}}. 
We contend that it is imperative to flexibly explore a larger pool of candidates to prioritize documents that present alternative evidence in diverse perspectives.

In this paper, we propose a new approach by employing Fine-grained Feedback with Reinforcement Retrieval, named FFRR, to enable more informed and accurate evidence retrieval for enhancing LLM-based fact checking on real-world news claims.
We first prompt LLM to generate intermediate questions from various perspectives of a claim ~\cite{pan2023factchecking,chen2023complex} to retrieve relevant documents. 
Then, FFRR collects fine-grained feedback from LLM on the retrieved documents as rewards at two levels:
1) Document-level rewards are used to refine the list of retrieved documents, such that individual documents that are more likely to contain evidence can have a better chance of getting selected through the exploration and exploitation process.
2) Question-level rewards are used to promote alternative evidence that cannot be surfaced by any single-angle inquiry, for which we optimize top-1 document ranking in each intermediate question.
With the two-level rewards combined, we optimize the policy model of FFRR based on policy gradient to select evidence documents autonomously.
Our contributions are three-fold:
\begin{itemize}[leftmargin=*]
\item This is the first work using fine-grained LLM feedback to reward policy optimization of reinforcement retrieval for black-box LLM-enabled fact checking on real-world news claims. 
\item We turn the sparse, non-retrieval-oriented claim-level supervision signals to fine-grained rewards on candidate documents and intermediate questions, which facilitates retrieval policy optimization, without adding any overhead on inference.
\item Results on two public news claim verification datasets demonstrate that FFRR outperforms strong LLM-enabled and non-LLM baselines by a large margin.\footnote{Code and prompts data are available at \url{https://github.com/jadeCurl/FFRR}.}
\end{itemize}

\section{Related Work}

Advanced LLMs have demonstrated impressive capabilities in generating human-like verbal reasoning~\cite{wei2022chain,wang2022self,zhou2022least}. 
Recent works~\cite{press2022measuring,yao2023react,jiang2023active,pan2023factchecking,zhao2023verify,zhang-gao-2023-towards,zeng2024justilm} find that combining LLM's reasoning capability with accessibility to external knowledge is helpful in factuality-related tasks, such as fact-checking on FEVER dataset~\cite{thorne2018fever}.
However, existing works overlook the optimization of retrieval models to better help LLMs. Compared to verifying the claims originated from Wikipedia facts in FEVER, verification of news claims is more realistic and complex, which needs more diverse and evidential information beyond LLM's memory~\cite{chen2023complex}.

Claim decomposition and question generation have been employed for claim verification~\cite{fan2020generating,pan2021zero,ousidhoum2022varifocal,chen-etal-2022-generating,chen2023complex}.
Typically, these methods analyze the claim to identify entities ~\cite{ousidhoum2022varifocal} and focal points~\cite{fan2020generating} to generate a series of corresponding questions, which are then answered via search engines to aid fact-checkers in making decisions. 
These models usually require a manual collection of claim-question datasets for training. 
Recently, \citet{chen2023complex} prompt a black-box LLM (i.e., \texttt{text-davinci-003}) with
in-context examples by carefully choosing four input-decomposition pairs from the human annotations of \citet{chen-etal-2022-generating} to form a few-shot
prompt.
We follow a similar approach to generate questions in our work.

Reinforcement learning (RL) has predominantly focused on treating LLMs as an RL agent and aligning them with human feedback (RLHF)~\cite{ouyang2022training,xu2022learning,bai2022training,wu2023fine} or feedback generated by other LLMs~\cite{bai2022constitutional}. Additional approaches diverge from this line of work by considering LLMs as black boxes  
and use RL to optimize models for choosing or generating suitable prompts for LLMs to enhance their outputs~\cite{lu2023dynamic,peng2023check}.
This is motivated by the fact that LLMs' massive parameter space makes it impractical for most users to fine-tune. 
Our FFRR method concentrates on optimizing the retrieval model by leveraging feedback from LLMs, targeting to improve retrieval results to enhance LLMs for news claim verification.

The REPLUG~\cite{shi2023replug} approach is closely related to us, which optimizes the retrieval model for open-domain QA through KLD minimization with LLM feedback on top-K documents. 
We, however, optimize retrieval based on RL framework for news fact checking, which can autonomously explore and exploit any documents that are potentially evidential to avoid missing important evidence.

\begin{figure*}[t]
\centering
\subfigure[Document level]{
\includegraphics[width=2.95in]{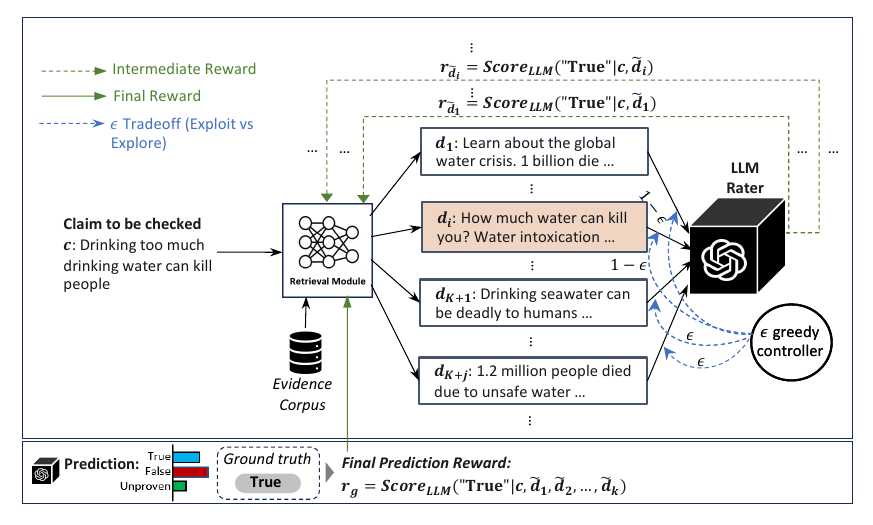}
\label{fig:model_0}
}
\subfigure[Question level]{
\includegraphics[width=3.15in]{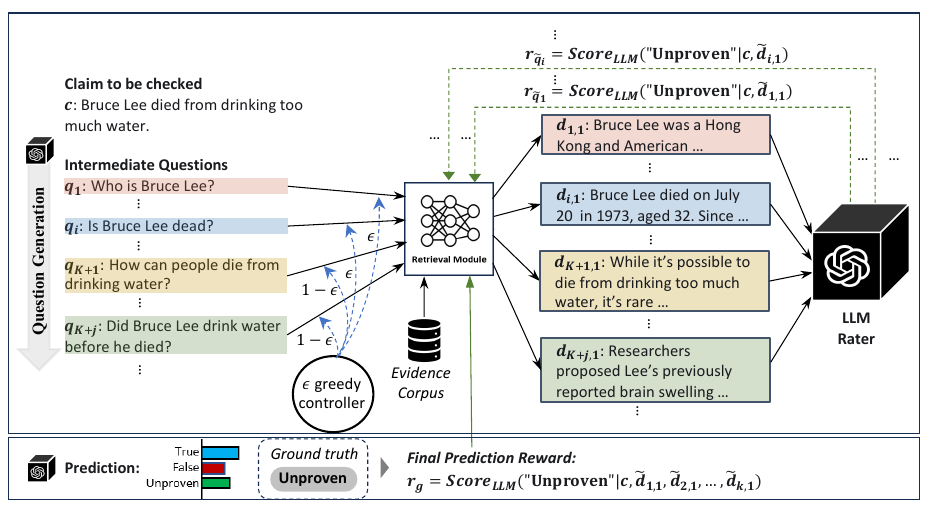} 
\label{fig:model_1}
}
\caption{FFRR policy models with different fine-grained rewards from LLM. In Figure~\ref{fig:model_0}, $d_i$ is shaded to indicate that it is the relevant document that potentially aids in verifying the truthfulness of the claim. }
\end{figure*}


\section{Problem Definition}

A news claim verification dataset consists of a set of news claims $\mathcal{C}=\{c\}$ which have been verified by public fact-checking services, and a large corpus of documents $\mathcal{D}=\{d_1,d_2,\cdots,d_{\mathcal{|D|}}\}$, which are pooled with the relevant documents gathered for each claim. It is worth to note that the documents from fact-checking sources and the raw documents published temporally after the corresponding fact-check reports should be strictly excluded from $\mathcal{D}$ to avoid potential leak of ground truth.
For any $c\in \mathcal{C}$, let $\mathcal{Q}=\{q_i\}_{i=1}^{|\mathcal{Q}|}$ denote a set of intermediate questions corresponding to $c$ obtained by claim decomposition (or question generation), and we use $\mathcal{D'} \subseteq \mathcal{D}$ to denote a set of documents retrieved by a query $q$ that is related to the claim $c$ (e.g., $q=c$ or $q\in \mathcal{Q}$).


Our task is to train a retrieval policy model $\pi_{\theta}$ with trainable parameters $\theta$ to output top-$K$ ($K\ll |\mathcal{D'}|$) documents for LLM to determine the veracity of a claim as one of the classes defined in benchmark datasets~\cite{wang2017liar,yang2022coarse}, such as true, mostly-true, half-true, barely-true, false, and pants-fire.
We use $\pi_{\theta}(d|q)$ to denote the retrieval probability of a document $d$ with respect to $q$. 

\section{Our Methodology}
\label{sec:model}
In this section, we will describe our FFRR model, starting from the basic document-level policy, followed by the question-level policy, and finally, a hybrid model that combines them.

\subsection{Document-level Retrieval Policy}
\label{sec:d}
We begin with a preliminary version of the FFRR model in Figure~\ref{fig:model_0}, where the claim is used directly as a query for retrieval, resulting in a ranking list of documents in the corpus.
The final goal of our policy model is to obtain top-$K$ documents that are most useful to LLM's final class prediction. 

To optimize the retrieval model, our policy aims to select a set of retrieved documents for gathering intermediate feedback on them from LLM.
Specifically, we use a dense retrieval model $\pi_\theta$ to obtain $\mathcal{D'}$ containing retrieved documents with respect to the given $c$.
We employ an $\epsilon$-greedy strategy to strike a balance between exploitation, i.e., choosing from the highly ranked documents such as top-$K$ in $\mathcal{D'}$, and exploration, i.e., selecting any documents in $\mathcal{D'}$.  
Thus, a document $\tilde{d}_i$ is sampled following one of the two document distributions according to
\begin{equation*}
\tilde{d}_i \sim
\begin{cases}
\text{Uniform}(\mathcal{D'}), & \text{explore with prob } \epsilon \\
\pi_\theta(c, K), & \text{exploit with prob } 1 - \epsilon
\end{cases}
\label{eq:epsilon}
\end{equation*}
where $\text{Uniform}(\mathcal{D'})$ is a uniform distribution over all retrieved documents, and $\pi_\theta(c, K)$ is the probability distribution of top-$K$ documents in $\mathcal{D'}$ retrieved by $\pi_{\theta}$, and $K$ and $\epsilon$ are hyper-parameters.
At each time step, we select a document followed by reward acquisition from LLM as described below.

\paragraph{Training Rewards.} During training, we consider the contribution of each selected document to making a \textit{correct} veracity judgment based on it as the \textit{document-level} intermediate reward. Specifically, for any sampled $\tilde{d}_i$, we compute a reward $r_{\tilde{d}_i}=\text{score}_{\text{LLM}}(y|c,\tilde{d}_i)$, where $\text{score}(.)$ is the score function based on LLM's output score of generating the \textit{ground-truth label token} $y$ given $c$ and $\tilde{d}_i$ as input, such as the probability or perplexity of the output token.

One might argue that the rewards generated by LLMs are likely biased by their preconceptions~\cite{tornberg2023chatgpt,shaikh-etal-2023-second}, potentially hampering retrieval performance. Why does it work provided that LLM might provide biased feedback due to such inherent limitation? Intuitively, the score function $r_{\tilde{d}_i}$ provides a natural facilitation mechanism to mitigate LLM bias on its veracity judgment, by telling the retriever what kind of evidence it returns will be helpful for LLM to make a right decision, which is governed by the \textit{ground-truth} veracity label. Take the claim ``Bruce Lee's death is due to drinking too much water'' with a ground-truth label ``Unproven'' as an example. If the LLM is strongly biased towards a different class, the reward score LLM gives to a retrieved document tends to be low since the posterior is ``Unproven''. In order to get a higher reward on the correct class, the retriever will be encouraged to select stronger evidence documents to counterbalance LLM’s biased preperceptions, so that LLM can face more and more concrete evidence returned that indicates the claim is unproven.

Note that the sampling process can terminate automatically. For the termination condition, we utilize the score distribution $\textbf{score}_{\text{LLM}}(\mathbf{y}|c,\tilde{d}_i)$ over \textit{all classes} $\mathbf{y}$ that LLM outputs given $c$ and a sampled $\tilde{d}_i$. The selection stops if $\text{argmax}_{y'\in \mathbf{y}} \sum_{i=1}^{\kappa} \textbf{score}(\mathbf{y}|c,\tilde{d}_i) = y$, where $y$ is the ground-truth label and $\kappa$ is a free number of \textit{actually sampled documents} upon termination.

Let us denote $k=\min(K,\kappa)$. Then we can propose the \textit{final} prediction reward as $r_g=\text{score}_{\text{LLM}}(y|c,\tilde{d}_1, \tilde{d}_2,\cdots,\tilde{d}_k)$, where $\tilde{d}_1,\tilde{d}_2,\cdots,\tilde{d}_k$ are the $k$ sampled documents being used by LLM to make the final prediction.  

\paragraph{Inference.} It is worth noting that the intermediate and final rewards are not needed once the retrieval model is trained. During inference, we just output the class token with the highest or the lowest score the LLM predicts depending on the specific metric used, e.g., probability or perplexity, by feeding top-$K$ retrieved documents into it. Hence, our approach does not introduce additional overhead to inference. It indeed costs more training time, but not significantly high, which is primarily caused by network traffic. We believe that a minor training overhead with much better effectiveness is still cost-effective. Such latency tends to decrease with the advancement of hardware infrastructure. 


\paragraph{Prompting LLM.} We prompt LLM to rate the selected documents for providing intermediate and final feedback for policy rewarding. We use the following prompt template for getting the rewards: 
\textit{``The following evidence is given: [DOC(, DOC,...)]. Among [LABEL SET], the claim [CLAIM] is classified as:''}. 
With the specific prompt text, LLM's API will return a score distribution of generating each class label as the next token. We then obtain the score corresponding to ground-truth label for calculating the reward. The illustrative example can be found in Table~\ref{tbl:demo0} in Appendix~\ref{app:demo}.


\subsection{Question-level Retrieval Policy}
\label{sec:q}
It has been recognized that claim verification often requires the consideration of multiple focal points to facilitate multi-hop reasoning~\cite{yao2023react,pan2023factchecking,chen2023complex}. To this end, we elicit LLM to decompose the claim $c$ into intermediate questions $\{q_i\}$ by employing in-context learning (ICL)  following~\cite{chen2023complex}. 
We instruct LLM to determine the number of questions automatically using demonstration examples. The detailed demonstrations are omitted here for space which can be found in Table~\ref{tbl:demo1} in Appendix~\ref{app:demo} .

Next, we optimize document retrieval for each intermediate question with a question-level policy, as shown in Figure~\ref{fig:model_1}. We focus on improving top-1 retrieval for each question to gather question-level rewards, thereby promoting the discovery of alternative evidence that is not surfaced through a single query alone. Hence, we sample top-1 documents across different questions. The question-level intermediate reward is thus calculated as $r_{\tilde{q}_i} = \text{score}_{\text{LLM}}(y|c,\tilde{d}_{i,1})$, where $\tilde{d}_{i,1}$ is the top-1 document retrieved by $\tilde{q}_i$ that is sampled out. 

Moreover, we want to promote the questions that should be better addressed by top-1 documents than those that are straightforward. For example, compared to ``Who is Bruce Lee'', questions like ``Did Bruce Lee drink water before he died'' and ``How can people die from drinking water'' are more complex and important, and it is more difficult to get the right top-1 documents for them. Thus, top-1 document sampling is intentionally biased towards the questions that receive lower $r_{\tilde{q}_i}$ score to give them a higher chance of getting selected. That said, we sort the questions reversely by the score and use the $\epsilon$-greedy method to select $k$ questions based on top-1 documents of all the questions. 

Finally, LLM utilizes the top-1 documents of $k$ selected intermediate questions to make the final prediction. The final prediction reward is given as $r_g=\text{score}_{\text{LLM}} (y|c,\tilde{d}_{1,1},\tilde{d}_{2,1},\cdots, \tilde{d}_{k,1})$, where the top-1 documents are from the $k$ sampled questions with intermediate rewards. The model is trained to better rank top-1 documents for those questions that need to be better addressed. During inference, we return the top-1 documents of $K$ questions that have the highest retrieval probability scores.

\subsection{Retrieval Policy with Hybrid Rewards}
\label{sec:h}

\begin{figure}[t!]
\centering
\includegraphics[width=3in]{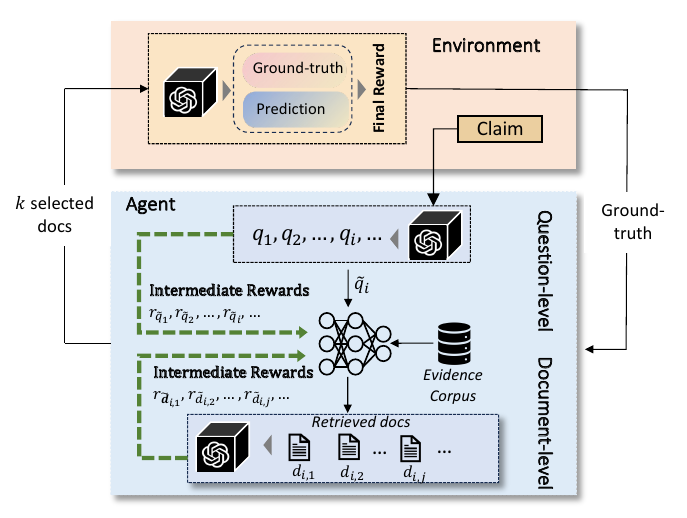} 
\caption{FFRR with two-level policies combined.}
\label{fig:model_2}
\end{figure}

We combine question- and document-level rewards to ensure that the model takes into account both breath and depth of evidence retrieval. The hybrid policy is shown in Figure~\ref{fig:model_2}, where each optimization step at question level is followed by a sequence of steps at document level and the two-level optimization proceeds alternately. 

We use two separate $\epsilon$-greedy controllers for the policy models of different levels. The document-level intermediate rewards is similar as described in \S~\ref{sec:d}, i.e., $r_{\tilde{d}_{i,j}}=\text{score}_{\text{LLM}}(y|c,\tilde{d}_{i,j})$ for any sampled intermediate question $\tilde{q}_i$. The question-level intermediate reward, and the final prediction reward are similar as described in \S~\ref{sec:q}.
In inference, using top-$K$ documents from all questions for final prediction is limited by LLM input constraints. So, we only take top-1 document for each question, mirroring the question-level strategy.

\section{Optimizing Retrieval Policy}

Following prior works~\cite{gao2022tevatron, karpukhin2020dense}, 
we use a dense retrieval model based on the dual-encoder architecture, where the encoder is a pre-trained RoBERTa~\cite{liu2019roberta} with parameter $\theta$, to map a query $q$ and a candidate document $d$ into vectors $\mathbf{h}_{\theta}(q)$ and $\mathbf{h}_{\theta}(d)$, respectively. 
The goal is to optimize the encoder $\mathbf{h}_{\theta}$ so that the query-document similarity serves as an effective retrieval function.

Specifically, we index all the documents in $\mathcal{D}$ offline using FAISS~\cite{johnson2019billion} which is a highly efficient open-source library for similarity search and clustering of dense vectors. 
The retrieval probability is defined as:
\begin{equation*}
\pi_\theta\left(d \mid q\right)= \frac{\exp(\mathbf{h}_{\theta}(q)^{\top} \cdot \mathbf{h}_{\theta}(d) / \tau)}{\sum_{d' \in \mathcal{D'}} \exp(\mathbf{h}_{\theta}(q)^{\top} \cdot \mathbf{h}_{\theta}(d') / \tau)},
\end{equation*}
where $\tau$ is the temperature hyper-parameter and $\mathcal{D'}$ is the set of retrieved documents.
In practice, we use a widely-used open-source toolkit Tevatron~\cite{gao2022tevatron} to build and optimize the dense retrieval model.

For optimizing retrieval policy $\pi_\theta$, with the goal of assisting the LLM, we adopt a policy gradient RL framework.
Each RL step corresponds to sampling a document provided to LLM for rewarding. 
We formulate the retriever-LLM interaction as a Markov Decision Process (MDP) described by:
\begin{itemize}[leftmargin=*]
    \item $S$ is an infinite set of states, which involves the claim, the documents selected in prior steps, and the current query to the retrieval model; 
    \item $A$ is a set of actions consisting of all the documents in $\mathcal{D}$; An action corresponds to selecting a candidate document $d$ returned by $\pi_\theta$.
    \item $R(s,a)$ is the reward received after taking action $a\in A$ in state $s\in S$. This reward incorporates both the fine-grained intermediate and final prediction rewards, as detailed in prior sections.
\end{itemize}

Our goal is to maximize $\mathbb{E}_{d \sim \pi_\theta\left(d \mid q\right)}[R(q,d)]$, the expected reward under the policy, with respect to the parameters using REINFORCE algorithm~\cite{williams1992simple}. The gradient of the expectation is approximated as:
\begin{equation*}
\begin{aligned}
&\nabla \mathbb{E}_{d \sim \pi_\theta\left(d \mid q\right)}\left[R(q,d)\right] 
\\& =\mathbb{E}_{d \sim \pi_\theta\left(d \mid q\right)} \left[\nabla_\theta \log \left(\pi_\theta\left(d \mid q\right)\right) R(q,d) \right]
\\& \approx \frac{1}{|\{(q,d)\}|} \sum_{(q,d)} \nabla_\theta \log \left(\pi_\theta\left(d \mid q\right)\right) R(q,d),
\end{aligned}
\end{equation*}
which is the mean of reward-weighted gradients over the selected documents. We update the retrieval policy so that model parameters can be learned via the rewards received by the documents that have been sampled. 

\section{Experiments and Results}

\subsection{Experimental Setup}

\subsubsection{Datasets.}
We conduct experiments on two public English news datasets referring to different fact-checking websites for ground-truth veracity labels: 1)  \textbf{RAWFC}~\cite{yang2022coarse} is based on claims and labels from Snopes in three classes (True/False/Half); 2) \textbf{LIAR-RAW}~\cite{yang2022coarse} contains claims and labels from PolitiFact in six classes (True/Mostly-true/Half-true/Barely-true/False/Pants-fire), based on the LIAR-PLUS dataset~\cite{alhindi-etal-2018-evidence}. 

We chose these two datasets because of their relatively lower ground-truth leakage than their counterparts like MultiFC~\cite{confemnlp2019Augenstein} which was found 17.2\% leaked evidence by our analysis using the leak detection method ~\cite{glockner2022missing}.
We used the raw documents in the two datasets collected by~\citet{yang2022coarse} for our corpus $\mathcal{D}$. These documents were retrieved from the web by excluding a range of fact-checking websites, and all of them were dated before publication of the fact-check reports that correspond to the claims. Using the detection method proposed by~\citet{glockner2022missing}, we further removed 4.7\% documents from RAWFC and 2.4\% documents from LIAR-RAW that are identified with leaks of ground truth.
Table~\ref{tbl_datasta} displays the statistics of the two datasets.\footnote{
We didn't choose FEVER~\cite{thorne2018fever} and its variants~\cite{jiang-etal-2020-hover,schuster-etal-2021-get} because 1) their claims are crafted based on Wikipedia articles, which don’t encapsulate the complexities inherent in real-world news-related assertions that we aim to verify; 2) Wikipedia data are commonly known as an important part of the training data used to pre-train many LLMs, which is likely to leak ground truth of claims. We also excluded the datasets consisting of only human-compiled evidence documents referenced in fact-check reports, such as WatClaimCheck~\cite{khan-etal-2022-watclaimcheck} and UKP-Snopes~\cite{hanselowski-etal-2019-richly}, since the evidence documents need to be retrieved out of noise.} 

Following~\citet{yang2022coarse}, we use macro-average precision ($P$), recall ($R$), and $F_1$ ($F_1=\frac{2 R P}{R+P}$) scores for evaluation. We utilize the supplied division of train-validate-test for both data sets. Specifically, they employed an 8/1/1 split ratio for the train/validation/test for the two datasets. Therefore, the corresponding number of samples in the RAWFC data for train/valid/test is 1,612/200/200, and the number of samples in the LIAR-RAW dataset is 10,065/1,274/1,251. 

\subsubsection{Implementation Details.}
We use GPT-3.5 API \texttt{text-davinci-003}\footnote{\url{https://platform.openai.com/docs/models/gpt-3-5}.} as a backbone LLM and generate outputs using greedy decoding. 
Following~\citet{wei2022chain}, we tune the shot number hyper-parameter within the set $\{1, 2, 4, 6, 8\}$ on the validation dataset. We observe that, although the model's performance experiences marginal gains with more than 2 shots, the increase in cost is substantial. Therefore, to balance cost and performance, we set the shot number of demonstration examples to 2. 

We use the open-source fine-tuning toolkit Tevatron~\cite{gao2022tevatron} to build the dense retrieval model. 
We pre-calculate the embeddings of documents $\mathcal{D}$ and create a FAISS index~\cite{johnson2019billion} to facilitate fast similarity search.
We retrieve a set of 20 documents as $\mathcal{D'}$ for each $q$ from the FAISS index, and set the temperature $\tau=1$. 
The selection of $\epsilon$ and $K$ is based on the validation set. We examine a range of $\epsilon$ from 0 to 1 and find that an $\epsilon$ value of 0.1 gives a good balance between exploration and exploitation. Specifically, when $\epsilon>0.1$, the model explores overly frequently, making it focus on random documents and hard to converge; when $\epsilon<0.1$, the model seems to always exploit, failing to select any new evidence. We adjust $K$ within the range of $[1,8]$ and fix it as 3 based on the observation that model performance generally declines when $K>3$.
We train the retriever using Adam optimizer~\cite{kingma2014adam} with a learning rate of 1$e$-5, a batch size of 4, and a warm-up ratio of 0.1. 
\begin{table}[t!]
\centering
\begin{tabular}{lrr}
\toprule

& \multicolumn{1}{r}{\textbf{RAWFC}}&\multicolumn{1}{r}{\textbf{LIAR-RAW}}\\

\midrule
Claim&2,012&12,590  \\
~~~\# true&695&2,021  \\
~~~\# mostly-true&-&2,439  \\
~~~\# half-true$^*$&671&2,592  \\
~~~\# barely-true &- &2,057  \\
~~~\# false&646&2,465 \\
~~~\# pants-fire&-&1,012\\
\hline
Avg. doc per claim&20.0&12.0\\
Avg. sentence per doc&7.1&5.5\\
\bottomrule

\end{tabular}
\caption{Datasets statistics after removing the documents with detected ground truth leaks. \# half-true$^*$ is also denoted as \# half in RAWFC.}
\label{tbl_datasta}

\end{table}

\begin{table*}[t]
\centering
\begin{tabular}{lllllll}
\toprule

\multirow{2}*{\textbf{Model}}& \multicolumn{3}{c}{\textbf{RAWFC}}&\multicolumn{3}{c}{\textbf{LIAR-RAW}}\\ 
\cmidrule(lr){2-4} \cmidrule(lr){5-7} 
&$P(\%)$ & $R(\%)$ & $F_1(\%)$&$P(\%)$ & $R(\%)$ & $F_1(\%)$\\\midrule
End-to-End Optim. SOTA~\cite{yang2022coarse}&52.8 & 51.0 & 51.9&29.0 & 28.9 & 28.9\\
\midrule
Direct Prompting~\cite{brown2020language}&48.5  &   48.5   &   48.5&29.1   &   25.1    &  27.0\\
Question-based Prompting&48.3 &     47.9 &     48.1 &27.1  &  24.5  &   25.7 \\
\midrule
ReAct~\cite{yao2023react}&51.2&48.5&49.8&33.2&29.0&31.0   \\
Verify-and-Edit~\cite{zhao2023verify}& 53.1    & 52.8   & 52.9&32.7&28.1&30.2\\
\midrule
REPLUG (d)~\cite{shi2023replug}&51.8&56.9&54.2&30.6&27.9&29.2\\
REPLUG (q)~\cite{shi2023replug} &52.4&56.0&54.1&30.0&28.7&29.3\\
REPLUG (d+q)~\cite{shi2023replug}&52.5&56.9&54.6&30.3&28.9&29.6\\
\midrule
FFRR-frozen (d)&53.6 & \textbf{57.9} &55.7 &30.7&29.5&30.1\\
FFRR-frozen (q) &51.9&54.2&53.0&29.9&27.3&28.5 \\
FFRR-frozen (d+q) &51.9&54.2&53.0&29.9&27.3&28.5 \\
FFRR (d)&55.5&56.2&55.8&33.0&29.8&31.3\\
FFRR (q)&54.2&56.0&55.1&32.8&28.4&30.4\\ 
FFRR (d+q)&\textbf{56.5}$^*$ & 57.4 & \textbf{57.0}$^*$ & \textbf{34.5}$^*$ & \textbf{32.6}$^*$ & \textbf{33.5}$^*$ \\
\bottomrule
\end{tabular}

\caption{Results of news claim verification on the two datasets where potential leaks are removed. Bold denotes the best performance. $^*$ means significantly better than the best baseline with $p<0.01$. LLMs involved in different models are based on GPT-3.5 (text-davinci-003).}
\label{tbl:main}
\end{table*}
\subsubsection{Compared Methods}
\paragraph{Baselines.}
1) \textit{End-to-End optimized SOTA}~\cite{yang2022coarse}, which is not LLM-based. Both its retrieval module and classifier are tunable.
2) \textit{Direct Prompting}~\cite{brown2020language}, which directly prompts the LLM to determine the class label of the claim. 
3) \textit{Question-based Prompting}, which asks the LLM to decompose the claim into questions and then predict the label by answering the questions.
4) \textit{ReAct}~\cite{yao2023react}, which originally combines LLM's reasoning ability and external Wikipedia API in an interleaved manner. We opt for Google Search API\footnote{\url{https://serpapi.com}. The choice aligns with previous work in news claim verification that predominantly employs search engines for retrieving relevant information~\cite{popat2018declare,ma2019sentence,yang2022explainable}.} instead of Wikipedia API since news claims typically require extensive and diverse information (e.g., blog posts, forum threads, and authoritative news) for verification.
5) \textit{Verify-and-Edit}~\cite{zhao2023verify}, which post edits CoT-style reasoning chains with external knowledge. Since it showed that using Google Search outperformed other retrieval methods on a fact-checking dataset, we adopt this setting as our baseline.
6) \textit{REPLUG}~\cite{shi2023replug}, which adapts the dense retrieval model to a black-box LLM by minimizing the KLD between the distributions of the likelihoods of retrieval and language model. We tailor REPLUG to our document-level (d), question-level (q), and hybrid (d+q) retrieval settings. For REPLUG (q), we optimize the KLD composed of retrieved top-1 documents from $K$ questions of each claim. For REPLUG (d+q), we optimize the KLD of top-$K$ documents of each question for all the $K$ questions of each claim, which are then combined with the KLD of top-1 documents of the $K$ associated questions.

\paragraph{FFRR Variants.}
1) \textit{FFRR-frozen}: Our FFRR with the retrieval model frozen. 
2) \textit{FFRR}: Our FFRR with tunable retrieval model as detailed in \S~\ref{sec:model}. 
Each of the two settings corresponds to the document-level (d), question-level (q), and hybrid (q+d) variants. Note that during inference,
both FFRR (q) and FFRR (d+q) use top-1 document for each question, as described in \S~\ref{sec:h}. Therefore, FFRR-frozen (d+q) and FFRR-frozen (q) are the same because retriever's parameters and retrieved documents are fixed, leading to the same set of top-1 documents of the questions between the two settings.

\subsection{Results of Verification}

Table~\ref{tbl:main} shows our results with the following findings:
\begin{itemize}[leftmargin=*]
\item \textbf{Simple zero-shot prompting methods already show competitive performance.} Direct Prompting lags behind by only 7.0\% in $F1$ score compared to the non-LM-based end-to-end optimized SOTA model. This suggests that LLM is promising in zero-shot news claim verification, thanks to LLM's ability acquired from its pre-training data, which embeds a substantial amount of factual information. However, we find that Question-based Prompting performs slightly worse. This might be attributed to the fact that Question-based Prompting demands more knowledge to answer intermediate questions, which LLM might not have, implying that more accurate information is needed from external sources.

\item \textbf{Retrieval benefits LLM.}
The performance improves when combining the LLM with methods for acquiring external knowledge from the web. ReAct and Verify-and-Edit integrate the multi-step reasoning capabilities of the LLM with an open-domain search engine to obtain relevant information, thus outperforming Direct Prompting and Question-based Prompting.
Interestingly, FFRR-frozen utilizing the frozen dense retrieval model clearly boosts the performance even higher on RAWFC and is just marginally lower than ReAct and Verify-and-Edit on LIAR-RAW.
This suggests the strong potential of combining small retrieval models with black-box LLMs for the task.

\item \textbf{Tuning the dense retrieval model with FFRR is superior.}
Our FFRR consistently provides improvements. Compared to FFRR-frozen, the FFRR (d), (q), and (d+q) demonstrate an increase of 1.7\%, 5.3\%, and 12.5\% on the $F1$ score, respectively.
REPLUG cannot compete with it because REPLUG is unable to flexibly explore and exploit all the retrieved documents in $\mathcal{D'}$ but sticks to its top-$K$ results. 
Furthermore, the claim decomposition of FFRR allows for searching evidence documents from various question perspectives. Despite this expanding the coverage, the questions as queries lead to more question-oriented evidence, which may not necessarily be helpful for the final prediction. This also account for the reason why FFRR (q) is slightly worse than FFRR (d), since the latter can flexibly explore the retrieved documents to the greater depth of the ranking list. When combining both document-level and question-level rewards, FFRR (d+q) enjoys the advantages of both, thereby outperforming all these baselines.
\end{itemize}

\subsection{Result Analysis}

\begin{figure}[t!]
\centering

\subfigure[RAWFC]{
\includegraphics[width=0.22\textwidth]{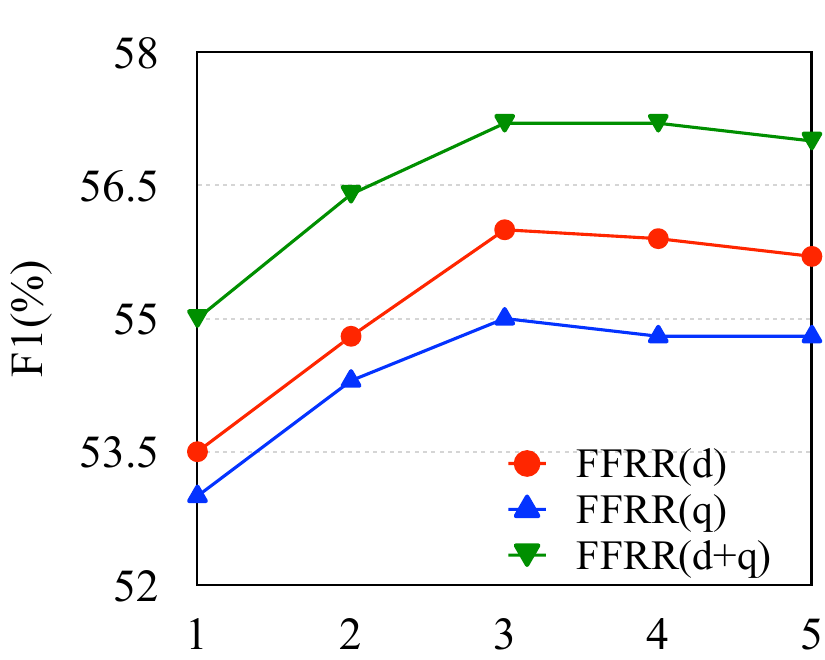}
}
\subfigure[LIAR-RAW]{
\includegraphics[width=0.22\textwidth]{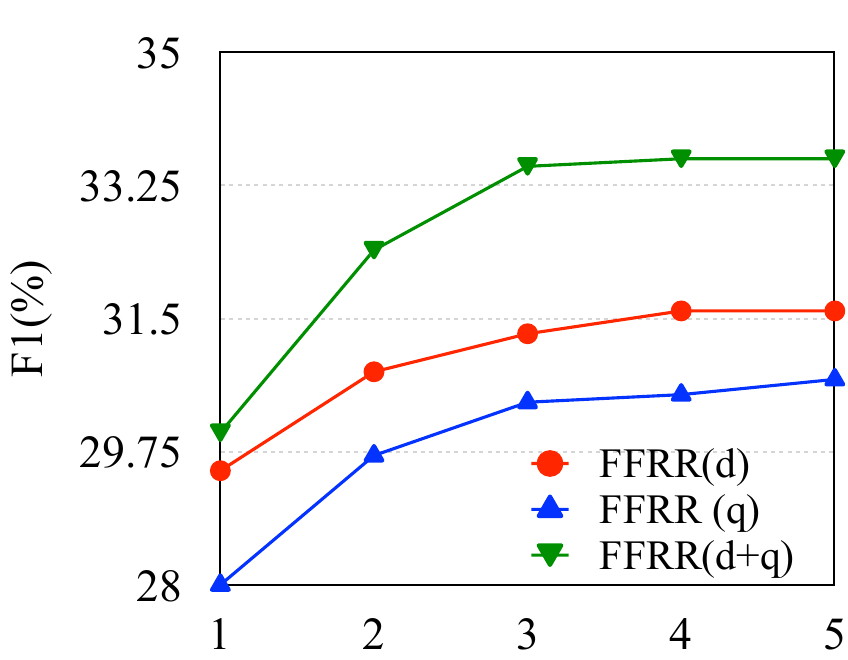}
}
\caption{Effect of number of documents $K$.}
\label{fig:k}
\end{figure}

\paragraph{Effect of the number of documents $K$.}
Firstly, we examine the effect of the number of documents $K$ by adjusting its value in the range from 1 to 5 for FFRR (d), FFRR (q), and FFRR (d+q) based on the test sets. As shown in Figure~\ref{fig:k}, the increase of $K$ improves the performance consistently when $K\le 3$. Our analysis indicates that while a too small $K$ (e.g., $K=1$) might still be able to find some documents needed to cover necessary evidence, a large $K$ (e.g., $K=5$) tends to include more irrelevant or noisy information generally leading to no further improvement. This suggests that the best option for guiding the LLM-based verification would be just using the top 3-4 documents.

\begin{figure}[t!]
\centering
\subfigure[RAWFC]{
\includegraphics[width=0.45\textwidth]{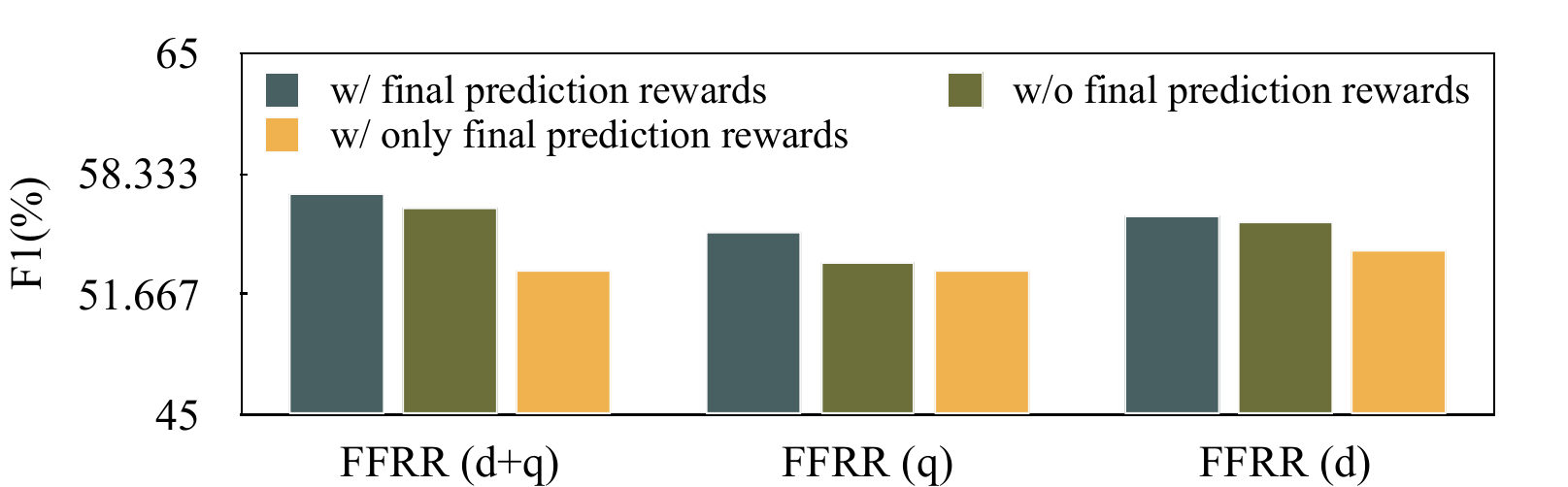}
\label{fig:fr_rawfc}
}
\subfigure[LIAR-RAW]{
\includegraphics[width=0.45\textwidth]{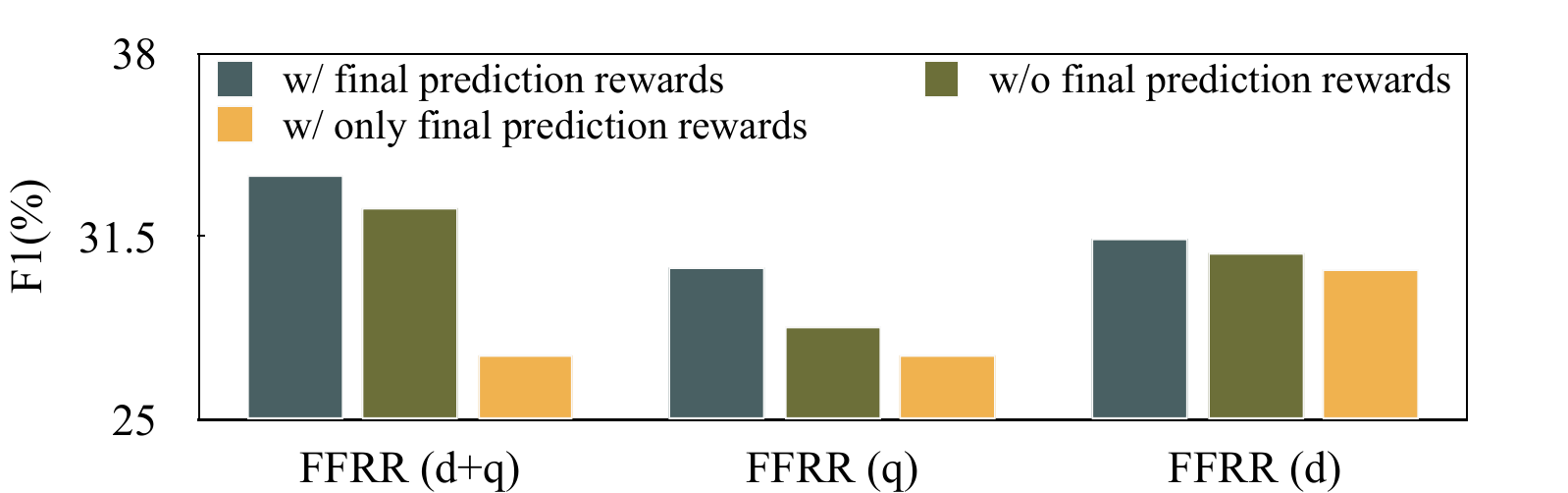} 
\label{fig:fr_liar}
}
\caption{Effect of final reward $r_g$.}
\label{fig:fr}
\end{figure}

\paragraph{Effect of final reward $r_g$.}
To analyze the impact of the final prediction reward of FFRR, we conduct an ablation analysis as shown in Figure~\ref{fig:fr}. 
For all three settings, i.e., d, q, and d+q, the performance of the model decreases on both datasets after removing the final prediction reward. This indicates that the final prediction reward is necessary and helpful, because it can measure the usefulness of the set of selected documents put altogether to influence LLM's judgement. However, if we only use the final reward for tuning, the performance is unsatisfactory. This confirms our hypothesis that optimizing the retrieval model with the sparse final feedback alone is insufficient, and thus our more fine-grained feedback scheme is necessary.

\paragraph{Error analysis of the retrieval process.}
\label{app:err}
To delve deeper into the FFRR model's performance in retrieval, we conduct a more detailed analysis.
Specifically, we randomly sample 50 instances that are predicted incorrectly by FFRR (i.e., failure cases) and manually identify four types of errors:
(1) \textit{Irrelevant questions:} this occurs when the generated intermediate questions do not pertain to or can not validate the claim;
(2) \textit{Insufficient Coverage:} the generated intermediate questions can not adequately cover the required information;
(3) \textit{Redundant questions:} the generated questions are repetitive or unnecessary for the verification;
and (4) \textit{Document Mismatch:} the documents retrieved can not answer the generated intermediate questions. In other words, there are clear mismatches or information gaps between the retrieved documents and the questions.

\begin{table}[t]
\centering
\begin{tabular}{lc}
\toprule
 Error Types&FFRR(d+q)\\
\hline
Irrelevant Questions & 16\%    \\
Insufficient Coverage& 34\%\\
 Redundant Questions& 14\% \\
 Document Mismatch & 48\%\\
\bottomrule
\end{tabular}
\caption{Distribution of errors based on 50 examples from RAWFC, where FFRR(d+q) gives incorrect verification results.}
\label{tab:error_distribution}
\end{table}

As shown in Table~\ref{tab:error_distribution}, we manually label the error types for these 50 instances. We clearly find that the most prevalent challenge lies in the document mismatch issue, which comprises a notable 48\% of the failure cases. This indicates that accurately retrieving evidential documents that can answer the raised questions is critical but desperately needs further improvement for providing necessary relevant context to the LLM. Following closely, the insufficient coverage issue, contributing to 34\% of the failures, suggests that sometimes specific nuances or details are not sufficiently obtained from the retrieval process. This can lead to situations where not all facets of a claim are adequately corroborated, resulting in incorrect predictions. On the brighter side, irrelevant and redundant questions appear to be less of a main concern. This gives us confidence that the LLM is generally adept at formulating intermediate questions that pertain directly to the given claim without being repetitive.


\section{Conclusion and Future Work}
In this paper, we study the reinforcement optimization of the retrieval model to retrieve evidence for promoting the capability of black-box LLM on fact verification of real-world news claims. We propose a method called FFRR utilizing Fine-grained Feedback for Reinforcement Retrieval, which adopts a two-level strategy to gather different rewards from the LLM based on retrieved documents to optimize the retrieval policy.
Evaluated on two public datasets, our LLM-based verification model leveraging FFRR significantly outperforms state-of-the-art LLM-enabled and non-LLM baselines.
In the future, we will develop a conversational news claim verification model based on LLMs that can receive fine-grained human feedback used to better align retrieval models with professional fact-checkers. Furthermore, we can also employ FFRR to optimize tunable retrieval models working in conjunction with LLMs on other NLP tasks.

\section{Limitations and Ethics Statement}
While our FFRR has demonstrated promising results, several limitations should be acknowledged. 

Firstly, although the find-grained reward is based on the final ground-truth label, our method greatly relies on the LLM's capabilities to assess the usefulness of selected documents to the prediction of the ground truth, as well as to make the final prediction. The performance of the LLM has a direct impact on our method's outcomes.
The LLM's inherent limitations, such as gaps in its understanding of specific contexts or language nuances, can influence the accuracy of our retrieval model and subsequent verification results. Similarly, any intrinsic bias in the LLM stemming from its training data can potentially skew the reward towards certain perspectives, thereby affecting the verification outcomes. We need further study on the influence of these issues on our task.

Secondly, the efficiency of training the reinforcement retrieval process could be handicapped by the frequent interactions for receiving feedback from the LLM. On the one hand, wse may need a systematic approach to optimize the feedback loop using more advanced exploration strategies to reduce the number of interactions needed. On the other hand, accelerating LLM inference can be achieved by integrating techniques such as non-autoregressive transformer~\cite{gu2017non,du2021order,du2022ngram} and speculative decoding~\cite{leviathan2023fast,du2024glide}.

Thirdly, the domain of this study is primarily focused on the verification of news claims only, which is just a portion of broader problem of misinformation. Our concentration on news claims can hopefully path for future studies, which could bed extended to check other forms of misinformation, such as fact verification~\cite{zhang2023ecenet,zeng-gao-2023-prompt} and infodemic surveillance~\cite{zhang2024predicting}.

The potential ethical concern associated with this research might be due to the generally low accuracy of news claim verification. The reliability of the system would be a primary concern when applied in the real world. Mislabeling false information or vice versa could have potential societal harm. 
Therefore, we advocate for the careful application of our model in real world. While our research contributes to the fight against misinformation, we believe the battle should also be supported by responsible journalism and informed public discourse.

\section*{Acknowledgement}
We thank the anonymous reviewers for their helpful comments during the review of this paper.

\bibliographystyle{lrec-coling2024-natbib}
\bibliography{lrec-coling2024-example}

\appendix
\section{Examples of Instruction Prompts}
\label{app:demo}
Table~\ref{tbl:demo0} and~\ref{tbl:demo1} display examples that serve as prompts for the LLM to make the final prediction and to generate intermediate questions, respectively.

\begin{table*}[t!]
\centering
\small
\begin{tabular}{p{15cm}}
\toprule
\rowcolor{gray!25}The following evidence is given:\textit{ [DOC(, DOC, ...)]}. Among \textit{[LABEL SET]}, the claim '''\textit{Emerson Moser, who was Crayola’s top crayon molder for almost 40 years, was colorblind.}''' can be classified as \textit{\ul{true}}.
\\
\rowcolor{gray!25}The following evidence is given:\textit{ [DOC(, DOC, ...)]}. Among \textit{[LABEL SET]}, the claim '''\textit{Bernie Sanders said 85 million Americans have no health insurance.}''' can be classified as \textit{\ul{half-true}.}
\\
\rowcolor{gray!25}The following evidence is given:\textit{ [DOC(, DOC, ...)]}. Among \textit{[LABEL SET]}, the claim '''\textit{Cheri Beasley “backs tax hikes — even on families making under \$75,000.}''' can be classified as \textit{\ul{False}.}
\\\\
\rowcolor{yellow!25}The following evidence is given:\textit{ [DOC(, DOC, ...)]}. Among \textit{[LABEL SET]}, the claim '''\textit{JAG charges Nancy Pelosi with treason and seditious conspiracy.}''' can be classified as \\
\rowcolor{green!25}\textit{\ul{False}.}\\
\bottomrule

\end{tabular}
\caption{An illustrative example of prompting LLM to make the final prediction. Examples used for demonstration are shaded in grey, with the claim undergoing verification highlighted in yellow and the LLM's prediction marked in green.}
\label{tbl:demo0}

\end{table*}

\begin{table*}[t!]
\centering
\small
\begin{tabular}{p{15cm}}
\toprule
\rowcolor{gray!25} \textbf{Claim:} \textit{Emerson Moser, who was Crayola’s top crayon molder for almost 40 years, was colorblind.} \\
\rowcolor{gray!25}To verify the claim, a fact-checker will go through a step-by-step process to ask and answer a series of questions relevant to its factuality. Here are the specific questions raised:\\
\rowcolor{gray!25}\textbf{Question:} Is there any official record or documentation indicating that Emerson Moser worked as a crayon molder at Crayola?\\
\rowcolor{gray!25}\textit{Question:} Are there any official records or documentation confirming Emerson Moser's length of employment at Crayola?\\
\rowcolor{gray!25}\textbf{Question:} Are there credible sources or publications that mention Emerson Moser as Crayola's top crayon molder?\\
\rowcolor{gray!25}\textbf{Question:} Are there any credible sources or records indicating that Emerson Moser was colorblind?\\
\rowcolor{gray!25}\textbf{Question:} Was Emerson Moser's colorblindness only confusing for certain colors?\\
\\
\rowcolor{gray!25}\textbf{Claim:} \textit{``Bernie Sanders said 85 million Americans have no health insurance.''}\\
\rowcolor{gray!25}To verify the claim, a fact-checker will go through a step-by-step process to ask and answer a series of questions relevant to its factuality. Here are the specific questions raised:\\
\rowcolor{gray!25}\textbf{Question:} How many Americans did Bernie Sanders claim had no health insurance?\\
\rowcolor{gray!25}\textbf{Question:} How did Bernie Sanders define ``no health insurance''?\\
\rowcolor{gray!25}\textbf{Question:} How many Americans were uninsured or under-insured according to the Commonwealth Fund survey?\\
\\
\rowcolor{gray!25}\textbf{Claim:} \textit{``Cheri Beasley “backs tax hikes — even on families making under \$75,000.''}\\
\rowcolor{gray!25}To verify the claim, a fact-checker will go through a step-by-step process to ask and answer a series of questions relevant to its factuality. Here are the specific questions raised:\\
\rowcolor{gray!25}\textbf{Question:} Does Cheri Beasley supports tax increases?\\
\rowcolor{gray!25}\textbf{Question:} Does the ad accurately link Beasley's position on student loan debt forgiveness with her stance on tax hikes for families making under \$75,000 per year?\\
\\
\rowcolor{yellow!25}\textbf{Claim:} \textit{``JAG charges Nancy Pelosi with treason and seditious conspiracy.''}\\
\rowcolor{yellow!25}To verify the claim, a fact-checker will go through a step-by-step process to ask and answer a series of questions relevant to its factuality. Here are the specific questions raised:\\
\rowcolor{green!25}\textbf{Question:} Is it true that JAG has made a claim or accusation against Nancy Pelosi?\\
\rowcolor{green!25}\textbf{Question:} Is it true that the specific charges or allegations made against Nancy Pelosi are treason and seditious conspiracy?\\
\bottomrule

\end{tabular}
\caption{An illustrative example of prompting LLM to generate intermediate questions. Examples used for demonstration are shaded in grey, with the claim undergoing verification highlighted in yellow and the LLM's generation marked in green.}
\label{tbl:demo1}

\end{table*}
\end{document}